\documentclass[letterpaper]{article} 
\usepackage[preprint]{aaai2027}  
\usepackage[hyphens]{url}  
\usepackage{graphicx} 
\usepackage{natbib}  
\usepackage{caption} 
\usepackage{algorithm}
\usepackage{algorithmic}
\usepackage{amssymb}
\usepackage{amsmath}
\usepackage{multirow}
\usepackage{newfloat}
\usepackage{listings}
\DeclareCaptionStyle{ruled}{labelfont=normalfont,labelsep=colon,strut=off} 
\floatstyle{ruled}
\newfloat{listing}{tb}{lst}{}
\floatname{listing}{Listing}

\usepackage{booktabs}

\title{EditFlow3D: Automated Local Editing of 3D Assets with Trajectory Preservation}
\author{
    Rui Nie\textsuperscript{\rm 1,\rm 3},
    Chuang Wang\textsuperscript{\rm 2}\thanks{Project lead.}, 
    Haitao Zhou\textsuperscript{\rm 1},
    Jiahe Song\textsuperscript{\rm 2},
    Buyu Li\textsuperscript{\rm 3},
    Sheng Wang\textsuperscript{\rm 3},
    Qian Yu\textsuperscript{\rm 1}\corresponding
}
\affiliations{
    \textsuperscript{\rm 1}Beihang University, 
    \textsuperscript{\rm 2}Shanghai Jiao Tong University,
    \textsuperscript{\rm 3}Bambu Lab \\
    qianyu@buaa.edu.cn
}

\begin{document}

\maketitle

\begin{figure*}[htbp]
    \centering
    \includegraphics[width=\linewidth]{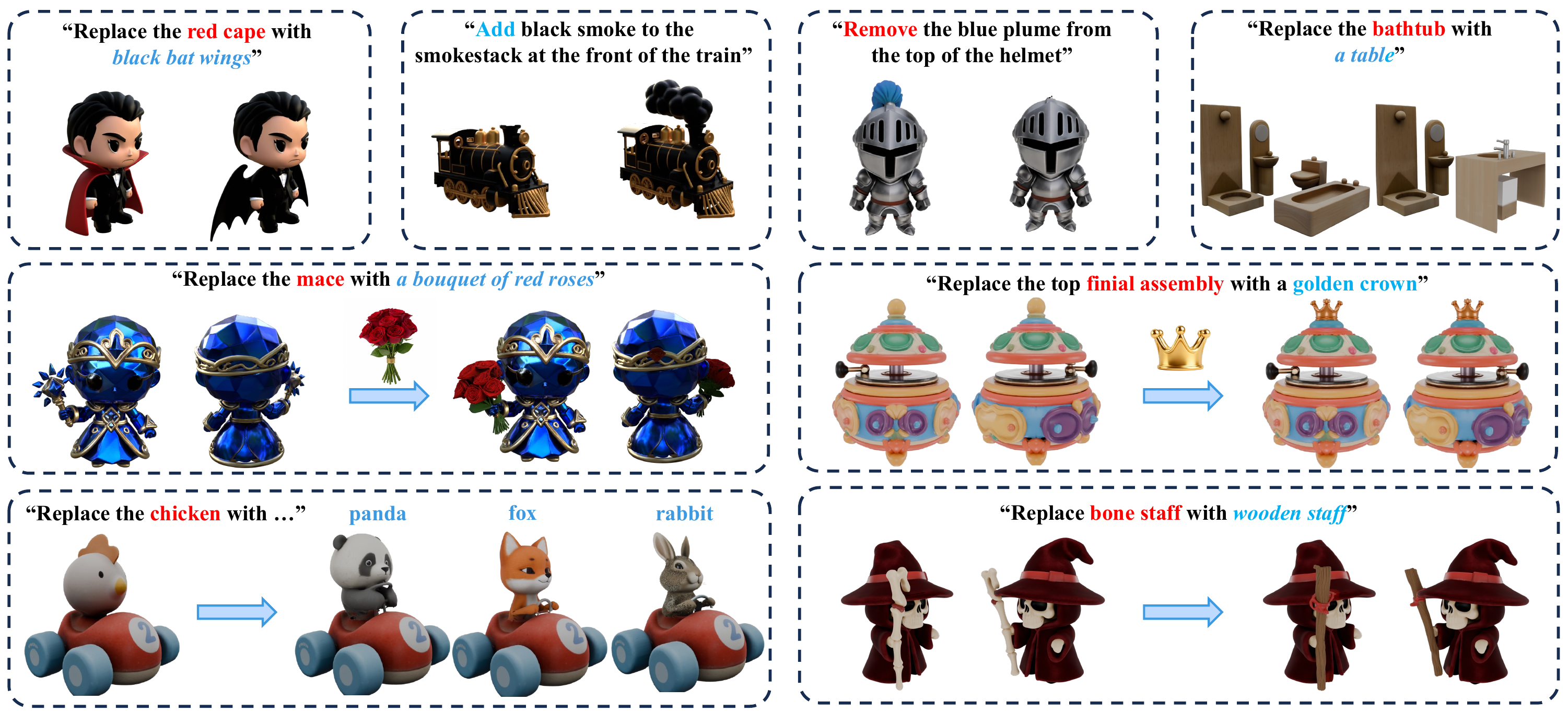}
    \caption{We present \textit{EditFlow3D}, a training-free and inversion-free framework for precise part-level editing of high-fidelity 3D assets. \textit{EditFlow3D} supports localized structural edits such as part replacement and removal while preserving the geometry and appearance of unedited regions.}
    \label{fig:teaser}
\end{figure*}

\begin{abstract}
Controllable local editing of 3D assets requires precise target localization and appropriate visual guidance. However, existing methods lack a simple yet accurate way to obtain 3D masks and struggle to achieve the desired edit while faithfully preserving the structure and appearance of non-target regions. To address these challenges, we present EditFlow3D, a training-free framework for local 3D editing. Given a source asset and an edit instruction, a VLM-driven workflow interprets the editing intent and automatically constructs a visual guidance image and a refined 3D editing mask, enabling localized editing in the native representation space of a pretrained 3D generative model. Specifically, mask-guided differential flow focuses the edit on the target region, while step-wise trajectory preservation maintains consistency between non-target regions and the source asset without directly replacing intermediate features. Since the existing Edit3D-Bench covers only a limited range of local editing categories, we further introduce EditFlow-Bench as a complementary benchmark encompassing a broader variety of structural and appearance edits, and evaluate EditFlow3D on both benchmarks. Quantitative results, qualitative comparisons, and a user study demonstrate that EditFlow3D achieves more accurate target-region editing and better preserves non-target regions than existing 3D editing methods.
\end{abstract}
\section{Introduction}
\label{sec:intro}

Recent advances in 3D generative models have made it possible to create detailed 3D assets from text or image conditions~\cite{hong2023lrm, shi2024mvdream, tang2025lgm, li2025triposg, hunyuan3d, step1x3d, craftsman3d}. Content creation, however, rarely ends with the initial generation. Users often need to modify an existing asset, such as replacing a component, removing unwanted geometry, or adding new content to a specified region. Unlike generating an asset from scratch, local editing must introduce the desired structural and appearance changes within the target region while retaining the materials, textures, and geometry of the remaining asset. This requires precise control over both the content and the spatial extent of the edit.

Existing methods explore 3D asset editing from different directions. Optimization- and inversion-based methods offer semantic flexibility, but their results depend on the quality of the optimized or inverted representation~\cite{sella2023vox}. Methods built on 2D image editing models~\cite{instantdit, preditor3d, pro3deditor, bar2025editp23} edit rendered views and then fuse or reconstruct them into a 3D asset, but may suffer from cross-view inconsistencies and reconstruction errors. More recent methods, including VoxHammer~\cite{voxhammer} and Nano3D~\cite{ye2025nano3d}, operate directly on native 3D representations or structured latent spaces, improving 3D structure awareness and cross-view consistency. However, they may still introduce unintended degradation in non-target regions, making it difficult to maintain consistency with the source asset during editing. Feedforward methods such as Easy3E~\cite{hu2026easy3e} and PartFlow~\cite{weng2026feedforward} learn editing transformations, but their generalization depends on the scale and quality of the training data. Despite these advances, local editing still requires balancing target modification against source preservation. A spatial mask can constrain the editing update, but it cannot prevent target conditioning from affecting non-target regions. Direct replacement with source features may strengthen preservation, but can disrupt cross-region consistency and introduce boundary artifacts. An effective editing mechanism must therefore localize the update while preserving non-target regions throughout editing.

Beyond the editing mechanism, reliable spatial controls are also essential for accurate local editing. Text and image inputs can describe the desired content but cannot explicitly specify the precise spatial extent of an edit in 3D. Recent work has therefore explored the automatic construction of multimodal editing controls. For example, Vinedresser3D~\cite{chi2026vinedresser3d} employs an VLM-based pipeline to interpret user requests and generate multimodal guidance, including a 3D mask, for target-region localization. While this approach demonstrates the potential of VLMs to make 3D editing more accessible, its automatically constructed 3D mask may remain imprecise, leading to inaccurate localization and unintended changes outside the target region. An effective automated editing pipeline should therefore generate not only appropriate visual guidance but also an accurate 3D mask that provides reliable spatial control for localized editing.

Motivated by these considerations, we propose \textbf{EditFlow3D}, a training-free framework for accurate local editing of 3D assets. EditFlow3D comprises two stages: automated editing-control construction and local 3D flow editing. In the first stage, a VLM-driven workflow interprets flexible user inputs, generates a visual guidance image, and constructs and refines a 3D editing mask, thereby automatically producing accurate visual and spatial controls. In the second stage, Differential Flow Guidance (DFG) derives the editing direction from the difference between source- and guidance-conditioned velocity predictions and uses the 3D mask to confine structural flow updates to the target region. Because visual conditioning is applied globally during subsequent feature generation, non-target regions may still drift from the source state. We therefore introduce Trajectory Preservation Guidance (TPG), which applies a source-state consistency constraint to non-target features at each generation step without directly replacing intermediate features, reducing unintended degradation and maintaining boundary coherence.

Furthermore, existing benchmarks cover only a limited range of local 3D editing tasks, making it difficult to comprehensively evaluate different methods across diverse structural and texture changes. To fill this evaluation gap, we construct \textbf{EditFlow-Bench}, comprising 100 source assets and 200 editing cases across five edit types spanning structural and texture changes. We evaluate different methods on EditFlow-Bench and the public Edit3D-Bench to provide a comprehensive comparison of their local editing capabilities.

Our contributions are summarized as follows:
\begin{itemize}
    \item We present \textbf{EditFlow3D}, a training-free framework that automatically converts flexible user inputs into accurate visual and spatial controls and performs precise local editing in the native representation space of a pretrained 3D generative model.

    \item We introduce a local 3D flow-editing algorithm that combines mask-guided differential flow with step-wise trajectory preservation. It restricts flow updates to the target region while suppressing non-target drift without hard replacement of intermediate features.

    \item We construct \textbf{EditFlow-Bench}, covering five edit types spanning structural and texture changes. Experiments on EditFlow-Bench and Edit3D-Bench using two TRELLIS-family backbones demonstrate that EditFlow3D achieves more accurate target-region editing and better non-target-region preservation than existing methods.

\end{itemize}

\begin{figure*}[t]
    \centering
    \includegraphics[width=\textwidth]{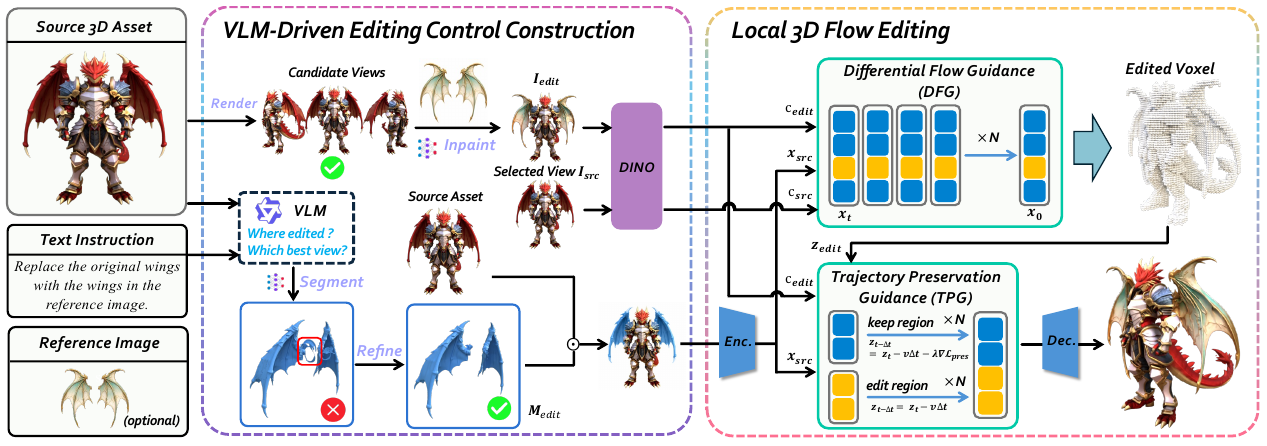}
    \caption{
    Overview of \textit{EditFlow3D}. A VLM-driven automated workflow converts a text instruction, an optional reference image, and multi-view renderings of a source 3D asset into a visual guidance image and a 3D editing mask. These controls guide masked flow updates in the target region, while trajectory preservation constrains the evolution of unedited regions.
    }
    \label{fig:arch}
\end{figure*}
\section{Related Work}
\label{sec:related}

\paragraph{Optimization- and view-based 3D editing.}
Early 3D editing methods optimize 3D representations using edited renderings or 2D generative priors. Vox-E~\cite{sella2023vox}, Instruct-NeRF2NeRF~\cite{haque2023instructnerf2nerf}, DreamEditor~\cite{zhuang2023dreameditor}, and GaussianEditor~\cite{wang2024gaussianeditor} apply text-guided edits to voxel grids, neural fields, meshes, or 3D Gaussians, but generally require per-asset iterative optimization. Another line edits rendered views and propagates or reconstructs the results in 3D. Instant3DiT~\cite{instantdit} performs multi-view inpainting followed by reconstruction, while PrEditor3D~\cite{preditor3d}, Pro3D-Editor~\cite{pro3deditor}, and EditP23~\cite{bar2025editp23} propagate edits across selected views. Despite benefiting from strong image editors, these methods remain susceptible to cross-view inconsistencies and reconstruction errors that may alter unedited content.

\paragraph{Native 3D and flow-based editing.}
Recent methods edit structured representations of pretrained 3D generative models. FlowEdit~\cite{kulikov2025flowedit} enables inversion-free image editing using the difference between source- and target-conditioned velocity fields. Nano3D~\cite{ye2025nano3d} extends this formulation to TRELLIS with region-aware voxel and latent-feature merging, while VoxHammer~\cite{voxhammer} inverts the source asset and replaces preservation-region features and attention tokens with cached source counterparts. Easy3E~\cite{hu2026easy3e} and PartFlow~\cite{weng2026feedforward} learn feedforward transformations for efficient editing. Although these methods improve efficiency and 3D consistency, their local behavior remains sensitive to region specification and source-preservation mechanisms. EditFlow3D applies conditional flow differences through a 3D mask and introduces trajectory preservation to softly constrain unedited regions without replacing intermediate features.

\paragraph{Editing-control construction.}
Local editing requires translating user requests into spatial and visual conditions. SAMPart3D~\cite{yang2024sampart3d} decomposes objects into semantic parts, PartField~\cite{liu2025partfield} learns hierarchical part-aware fields, and P3-SAM~\cite{ma2025p3} supports point-prompted native 3D segmentation. Part-X-MLLM~\cite{wang2025part} represents part grounding and editing commands through structured multimodal outputs. More closely related, Vinedresser3D~\cite{chi2026vinedresser3d} uses an MLLM to interpret instructions, select a view, generate visual guidance, and identify the region before inversion-based native 3D editing. EditFlow3D similarly automates visual and spatial control construction, while coupling the guidance image and refined 3D mask with masked flow updates and trajectory preservation in the native sparse representation.

\section{Method}
\label{sec:method}

\subsection{Overview}

As illustrated in Figure~\ref{fig:arch}, \textbf{EditFlow3D} consists of two stages: VLM-driven automated construction of editing controls and local 3D flow editing. Given a text editing instruction, an optional reference image, and a source 3D asset, the first stage jointly analyzes the user inputs and multi-view renderings of the source asset and invokes specialized tools to generate a visual guidance image and a 3D editing mask.

The second stage performs the requested edit in the native representation space of a pretrained TRELLIS-family model. TRELLIS-family models follow a hierarchical coarse-to-fine generation process: they first generate a sparse voxel structure that determines the occupied regions of an object and then progressively recover fine-grained geometry and texture representations by denoising the latent features associated with active voxels. Following this process, Differential Flow Guidance (DFG) operates during sparse voxel generation, using visual guidance to determine the editing direction and the 3D mask to confine structural updates to the target region. 
During subsequent feature generation, Trajectory Preservation Guidance (TPG) constrains non-target feature trajectories toward the source state while allowing target features to evolve under the visual editing condition, thereby reducing non-target drift and maintaining boundary coherence.

\subsection{VLM-Driven Editing Control Construction}

\subsubsection{View Selection and Guidance Image Generation}

To construct visual guidance for local 3D editing, we first render the source asset from a fixed set of predefined viewpoints, obtaining a fixed number of multi-view images. Given these rendered images, the text editing instruction, and an optional reference image, the VLM selects the view most suitable for the current editing task. The selected view should clearly expose the target region while retaining sufficient context of the source asset. We denote the selected source view as $I_{\mathrm{src}}$.

The VLM then invokes an image editing model to edit $I_{\mathrm{src}}$ according to the text instruction and the optional reference image, producing a guidance image $I_{\mathrm{edit}}$ that explicitly depicts the desired modification. Finally, $I_{\mathrm{src}}$ and $I_{\mathrm{edit}}$ serve as the source visual condition $c_{\mathrm{src}}$ and the target visual condition $c_{\mathrm{edit}}$, respectively, for subsequent local 3D flow editing.

\subsubsection{3D Editing Mask Construction}

The 3D editing mask is used to define the spatial extent of subsequent flow updates. The more accurately it delineates the intended editing region, the more effectively the method can modify the target while preserving non-target content.  Given the source asset, we first use P3-SAM~\cite{ma2025p3} to decompose it into parts. Operating natively in 3D, P3-SAM provides robust part recognition for complex assets and produces a relatively complete initial decomposition. We render its outputs as multi-view color-coded segmentation maps, in which each part is assigned a unique color and ID. Based on the text editing instruction, the VLM selects the corresponding part IDs, whose union forms the initial mask $M^{(0)}$.

Starting from $M^{(0)}$, we refine the geometric boundary of the target region. Since P3-SAM may produce region leakage near part junctions, we use the hierarchical partitions of PartField~\cite{liu2025partfield} and the multi-granularity partitions of S2AM3D~\cite{su2026s2am3d} to divide these ambiguous boundary regions into smaller candidate parts. To compare the outputs of different models, we map all segmentation results onto a common set of surface points sampled from the source asset. Let $S_k^{(m)}$ denote the $k$-th candidate part produced by auxiliary model $m$, and let $g_k^{(m)}$ denote its granularity. We measure its agreement with $M^{(0)}$ using overlap purity:

\begin{equation}
\begin{aligned}
P_k^{(m)}
&=
\frac{A\!\left(S_k^{(m)}\cap M^{(0)}\right)}
     {A\!\left(S_k^{(m)}\right)},
\end{aligned}
\label{eq:mask_matching}
\end{equation}

\noindent where $A(\cdot)$ denotes the surface area of a region. For each granularity $g$, we retain the purity-qualified candidate pool:

\begin{equation}
\mathcal Q_{m,g}=\left\{k\mid g_k^{(m)}=g,\;P_k^{(m)}\geq\tau_P\right\},
\end{equation}

\noindent where $\tau_P$ is a purity-threshold hyperparameter that controls the minimum overlap required for a candidate to enter the pool. Starting from an empty union, we then greedily select from $\mathcal Q_{m,g}$ the candidate that yields the largest positive improvement in the area-weighted $F_1$ score with respect to $M^{(0)}$, and add it to the union. The search stops when no remaining candidate improves the score or when the maximum number of selected parts is reached.



We further filter and fuse the candidates according to their geometric connectivity with the target region and surface-normal consistency. This retains local regions consistent with the target part while rejecting erroneous segments near its boundary, producing the refined surface mask $M_{\mathrm{ref}}$. Further details of candidate matching and boundary refinement are provided in the Appendix.A.

Finally, we map $M_{\mathrm{ref}}$ to the native sparse coordinates of the 3D generative model. For deletion, the mask covers the target region with a small boundary margin. For modification and replacement, it is expanded within a bounded neighborhood to accommodate potential structural changes. For addition, unoccupied locations outside the source asset are treated as editable regions. This produces the final editing mask $M_{\mathrm{edit}}$, whose complement $M_{\mathrm{keep}}$ constrains the unedited region.

\subsection{Local 3D Flow Editing}

\subsubsection{Differential Flow Guidance}
\label{sec:masked_flow_update}

After obtaining the edited guidance image and the 3D editing mask, we modify the source asset within the native representation space of a pretrained 3D generative model. To determine the editing direction without inverting the source asset, we follow the differential-flow principle of FlowEdit~\cite{kulikov2025flowedit}. We compare the velocity fields predicted under the source and target visual conditions and use their difference as the editing direction from the source asset toward the desired result. The 3D editing mask spatially constrains this differential velocity, restricting the flow update to the target region. We refer to this process as differential flow guidance (DFG).

Let $\mathbf{x}_{\mathrm{src}}$ denote the latent representation obtained by encoding the source 3D asset with the pretrained 3D encoder. We initialize the editable state as

\begin{equation}
    \mathbf{z}_{\mathrm{edit}}
    =
    \mathbf{x}_{\mathrm{src}}.
\end{equation}

At time step $t$, we construct a noisy source state following the linear path of rectified flow:

\begin{equation}
    \mathbf{z}_{\mathrm{src},t}
    =
    (1-t)\mathbf{x}_{\mathrm{src}}
    +
    t\boldsymbol{\epsilon},
    \qquad
    \boldsymbol{\epsilon}
    \sim
    \mathcal{N}(\mathbf{0},\mathbf{I}).
\end{equation}

The current editing displacement relative to the source asset is then transferred to the same noise level, yielding the corresponding target state:

\begin{equation}
    \mathbf{z}_{\mathrm{tgt},t}
    =
    \mathbf{z}_{\mathrm{src},t}
    +
    \left(
    \mathbf{z}_{\mathrm{edit}}
    -
    \mathbf{x}_{\mathrm{src}}
    \right).
\end{equation}

The flow editor estimates the editing direction by comparing the velocity predictions under $c_{\mathrm{src}}$ and $c_{\mathrm{edit}}$:

\begin{equation}
    \Delta v_t
    =
    \mathbb{E}_{\boldsymbol{\epsilon}}
    \left[
    v_{\theta}
    \left(
    \mathbf{z}_{\mathrm{tgt},t},
    t,
    c_{\mathrm{edit}}
    \right)
    -
    v_{\theta}
    \left(
    \mathbf{z}_{\mathrm{src},t},
    t,
    c_{\mathrm{src}}
    \right)
    \right].
\end{equation}

The first term describes the generation direction of the current state under the target visual condition, while the second describes that of the source asset under the source visual condition. Their difference therefore provides a differential velocity toward the desired editing result.

We map the 3D editing mask to the native representation space of the generative model and denote it by $\mathcal{S}_{\mathrm{edit}}$. The editable state is updated as

\begin{equation}
    \mathbf{z}_{\mathrm{edit}}
    \leftarrow
    \mathbf{z}_{\mathrm{edit}}
    +
    \gamma\Delta t\,
    \mathcal{S}_{\mathrm{edit}}
    \odot
    \Delta v_t,
\end{equation}

\noindent where $\gamma$ controls the editing strength and $\Delta t$ denotes the flow step size. The guidance image specifies the desired structural and appearance changes through the differential velocity, while the 3D editing mask restricts the corresponding flow update to the target region.

\subsubsection{Trajectory Preservation Guidance}
\label{sec:trajectory_preservation}

\begin{figure*}[h!]
    \centering
    \includegraphics[width=\linewidth]{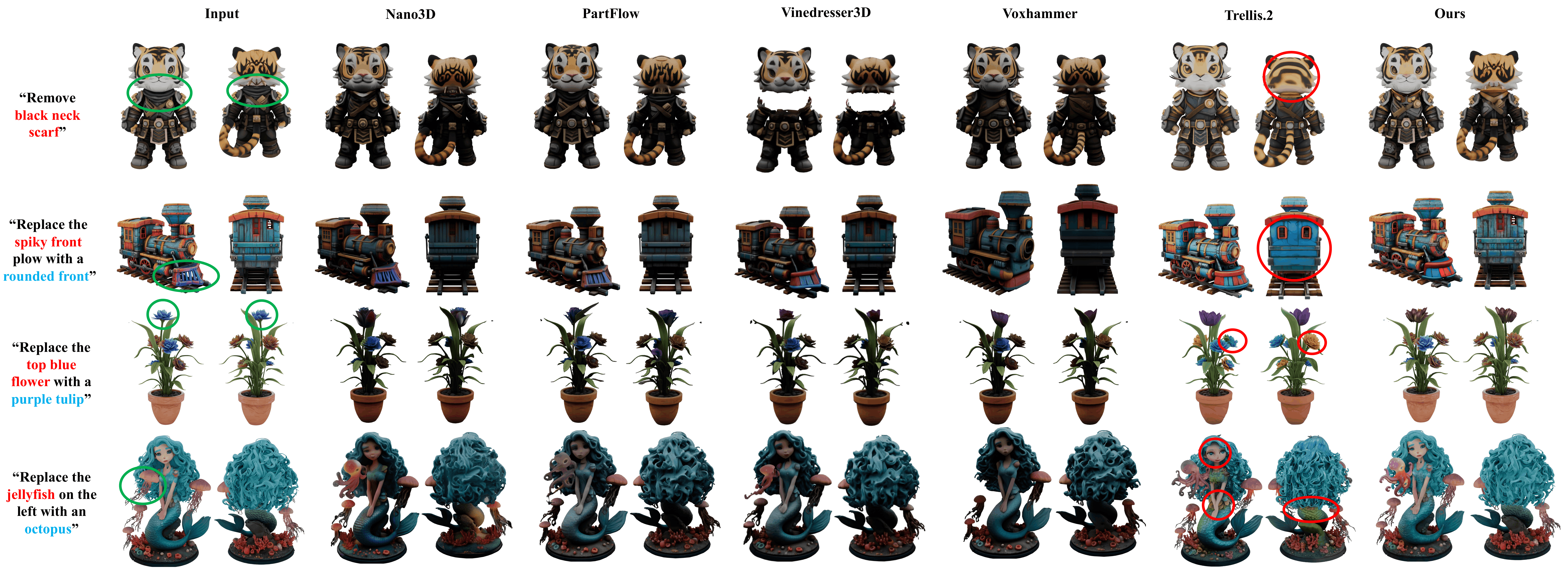}
    \caption{
        Qualitative comparison on representative local 3D part-editing cases.
        All editing methods use the same source asset and edited guidance image for each case.
        EditFlow3D uses TRELLIS.2 and better localizes the requested changes while preserving non-target geometry and appearance.
    }
    \label{fig:show_case}
\end{figure*}

Following the mask-guided structural update by DFG, the edited sparse voxel structure is passed to the subsequent conditional flow stage to generate geometry and texture features. Although DFG localizes the structural modification, the target visual condition is applied globally during feature generation and may still cause non-target features to drift from the source asset. We therefore apply trajectory preservation guidance (TPG) at each feature-generation step to keep non-target features close to their encoded source states while allowing target features to evolve under the editing condition. Unlike direct feature replacement, TPG enforces source consistency along the generation trajectory, thereby reducing unintended degradation and maintaining boundary coherence.

Under the target visual condition, we estimate the terminal clean state according to the linear trajectory of rectified flow:

\begin{equation}
    \hat{\mathbf{x}}_{0,t}
    =
    \mathbf{z}_{\mathrm{tgt},t}
    -
    t\,
    v_{\theta}
    \left(
    \mathbf{z}_{\mathrm{tgt},t},
    t,
    c_{\mathrm{edit}}
    \right).
\end{equation}

Let $\mathcal{S}_{\mathrm{keep}}=1-\mathcal{S}_{\mathrm{edit}}$. We define the preservation loss by comparing the predicted clean state with the source state within the preservation region:

\begin{equation}
    \mathcal{L}_{\mathrm{pres}}
    =
    \left\|
    \mathcal{S}_{\mathrm{keep}}
    \odot
    \left(
    \hat{\mathbf{x}}_{0,t}
    -
    \mathbf{x}_{\mathrm{src}}
    \right)
    \right\|_2^2.
\end{equation}

After the mask-guided differential flow update, we softly correct the editable state using the gradient of the preservation loss:

\begin{equation}
    \mathbf{z}_{\mathrm{edit}}
    \leftarrow
    \mathbf{z}_{\mathrm{edit}}
    -
    \lambda\,
    \left(
    \nabla_{\mathbf{z}_{\mathrm{edit}}}
    \mathcal{L}_{\mathrm{pres}}
    \right),
\end{equation}

\noindent where $\lambda$ controls the preservation strength. This encourages the unedited regions to remain close to the source trajectory while allowing the target region to evolve according to the edited guidance image, thereby reducing unintended changes and discontinuities near the editing boundary.


\section{Experiments}
\label{sec:exp}

\subsection{Implementation Details}
\label{sec:imp}

We implement EditFlow3D primarily on TRELLIS.2~\cite{xiang2025native}, which serves as our default backbone due to its higher-quality 3D generation. Since most existing baselines are built on TRELLIS~\cite{trellis}, we additionally implement a TRELLIS-based variant for fair comparison. For the VLM-driven editing control construction, we use Qwen3-VL-8B~\cite{qwen3vl2025} as the multimodal reasoning model, and generate edited guidance images using FLUX.1 Kontext [dev]~\cite{labs2025flux1kontextflowmatching}. For DFG, we use $25$ flow steps, set both the source- and target-conditioned classifier-free guidance scales to $5.0$, and average the differential velocity over four noise samples at each step. For TPG, we use $12$ flow steps and set the preservation strength to $\lambda=12$. All experiments are conducted on a single NVIDIA A800 GPU with $80$ GB of memory. Given an editing mask, the complete 3D editing process, including DFG, TPG, and final asset generation, takes less than one minute per asset. Unless otherwise specified, qualitative results use TRELLIS.2, while quantitative results are reported using both backbones. 



\paragraph{Datasets.}
We evaluate EditFlow3D on the public Edit3D-Bench and our newly constructed EditFlow-Bench, which contains $100$ source 3D assets: $30$ real-scanned assets from Google Scanned Objects (GSO)~\cite{downs2022scanned}, $30$ artist-created assets corresponding to real-world products from Amazon Berkeley Objects (ABO)~\cite{Collins_2022_ABO}, and $20$ assets generated by each of TRELLIS.2 and Hunyuan3D 2.1~\cite{hunyuan3d-2.1}. Each asset is associated with two distinct editing instructions, resulting in $200$ editing cases evenly distributed across five operations: addition, deletion, replacement, geometry modification, and appearance modification, with $40$ cases per operation. For each case, the 3D editing mask is automatically constructed using our automated control-construction workflow and subsequently manually inspected to ensure its correctness. The editing instructions and guidance images are generated using Gemini 3 Pro~\cite{gemini3pro2026} and FLUX.1 Kontext [dev], respectively. Additional construction and evaluation details are provided in the Appendix.B.

\begin{table*}[t]
\centering
{\small
\setlength{\tabcolsep}{2.5pt}
\begin{tabular}{l*{14}{c}}
\toprule
\textbf{Method}
& \multicolumn{7}{c}{\textbf{EditFlow-Bench (Automatic Mask)}}
& \multicolumn{7}{c}{\textbf{Edit3D-Bench (Oracle Mask)}} \\
\cmidrule(lr){2-8}
\cmidrule(lr){9-15}

& Sim.$\uparrow$
& $\mathrm{CD}_{k}\downarrow$
& PSNR$\uparrow$
& SSIM$\uparrow$
& LPIPS$\downarrow$
& DINO-I$\uparrow$
& FID$\downarrow$
& Sim.$\uparrow$
& $\mathrm{CD}_{k}\downarrow$
& PSNR$\uparrow$
& SSIM$\uparrow$
& LPIPS$\downarrow$
& DINO-I$\uparrow$
& FID$\downarrow$ \\
\midrule

Nano3D
& 0.275 & 0.221 & 25.42 & 0.826 & 0.153 & 0.883 & 74.50
& 0.281 & 0.007 & 30.52 & 0.967 & 0.026 & 0.930 & 23.12 \\

PartFlow
& 0.287 & 0.229 & 26.18 & 0.833 & 0.149 & 0.887 & 72.30
& 0.286 & 0.010 & 29.20 & 0.959 & 0.032 & 0.929 & 23.45 \\

Vinedresser3D
& 0.292 & 0.207 & 27.11 & 0.839 & 0.143 & 0.891 & 69.84
& 0.294 & 0.018 & 28.68 & 0.957 & 0.052 & 0.923 & 31.44 \\

VoxHammer
& 0.284 & 0.192 & 24.81 & 0.842 & 0.136 & 0.881 & 71.35
& 0.291 & 0.014 & 27.63 & 0.952 & 0.043 & 0.922 & 28.79 \\

TRELLIS
& 0.292 & 0.251 & 25.84 & 0.796 & 0.189 & 0.875 & 75.12
& 0.293 & 0.031 & 21.73 & 0.906 & 0.127 & 0.852 & 37.78 \\

TRELLIS.2
& \underline{0.314} & 0.235 & 27.62 & 0.814 & 0.174 & 0.895 & 68.27
& \underline{0.312} & 0.056 & 22.65 & 0.911 & 0.120 & 0.863 & 35.06 \\

\midrule

\textbf{Ours} (TRELLIS)
& 0.299
& \underline{0.158}
& \underline{34.76}
& \underline{0.892}
& \underline{0.091}
& \underline{0.923}
& \underline{49.20}
& 0.297
& \underline{0.006}
& \underline{36.84}
& \underline{0.986}
& \underline{0.023}
& \underline{0.939}
& \underline{21.46} \\

\textbf{Ours} (TRELLIS.2)
& \textbf{0.318}
& \textbf{0.144}
& \textbf{36.53}
& \textbf{0.904}
& \textbf{0.080}
& \textbf{0.934}
& \textbf{43.51}
& \textbf{0.315}
& \textbf{0.005}
& \textbf{39.28}
& \textbf{0.992}
& \textbf{0.020}
& \textbf{0.945}
& \textbf{19.87} \\

\bottomrule
\end{tabular}
}
\caption{
Quantitative comparison on EditFlow-Bench and the public
Edit3D-Bench. Bold and underlined values indicate the best and
second-best results, respectively.
}
\label{tab:quantitative_comparison}
\end{table*}

\paragraph{Baselines.}
We group the compared approaches into 3D editing baselines and direct reconstruction baselines. The editing baselines include Nano3D~\cite{ye2025nano3d}, PartFlow~\cite{weng2026feedforward}, VoxHammer~\cite{voxhammer}, and Vinedresser3D~\cite{chi2026vinedresser3d}. They are evaluated using official implementations and recommended configurations. We standardize the inputs by providing the same source 3D asset and edited guidance image to the editing methods. For mask-based methods, we follow benchmark-specific protocols: all such methods use the benchmark-provided oracle masks on Edit3D-Bench and the same automatically constructed masks from our control-construction workflow on EditFlow-Bench. We report direct reconstruction with TRELLIS and TRELLIS.2. These baselines do not edit the source 3D representation; instead, they reconstruct the edited guidance image using the corresponding generative model.

\paragraph{Evaluation Metrics.} We use CLIP-based Similarity (Sim.) to measure semantic alignment between the editing instruction and multi-view renderings of the edited region. Within the preservation region, $\mathrm{CD}_{keep}$ compares the source and edited geometry, while PSNR, SSIM, and LPIPS compare the source and edited renderings under the projected preservation masks. DINO-I~\cite{dinov2} compares the full source and edited renderings to measure global image-level consistency, and FID~\cite{fid} measures the distribution-level quality of the rendered results.

\begin{table}[t]
\centering
\resizebox{\columnwidth}{!}{%
\begin{tabular}{lccccccc}
\toprule
\textbf{Variant}
& Sim.$\uparrow$
& $\mathrm{CD}_{keep}\downarrow$
& PSNR$\uparrow$
& SSIM$\uparrow$
& LPIPS$\downarrow$
& DINO-I$\uparrow$
& FID$\downarrow$ \\
\midrule

Full
& \textbf{0.318}
& \textbf{0.144}
& \textbf{36.53}
& \textbf{0.904}
& \textbf{0.080}
& \textbf{0.934}
& \textbf{43.51} \\

Fixed View
& 0.308
& 0.167
& 33.84
& 0.892
& 0.098
& 0.923
& 49.86 \\

w/o mask Ref.
& 0.300
& 0.186
& 32.71
& 0.881
& 0.112
& 0.915
& 53.42 \\

w/o DFG
& 0.287
& 0.213
& 29.86
& 0.858
& 0.136
& 0.897
& 61.72 \\

w/o TPG ($\lambda=0$) 
& 0.306
& 0.194
& 27.95
& 0.846
& 0.158
& 0.889
& 67.84 \\

\bottomrule
\end{tabular}%
}
\caption{
Quantitative ablation on EditFlow-Bench. Each variant removes or replaces one control-construction or flow-editing component of the full method. “Fixed View” uses the predefined canonical view, while “w/o mask Ref.” directly uses the raw mask predicted by P3-SAM without refinement.
}
\label{tab:ablation_edit3d}
\end{table}

\subsection{Quantitative Evaluation}

As shown in Table~\ref{tab:quantitative_comparison}, the 3D editing baselines exhibit a trade-off between target-region alignment and non-target-region preservation, with no method performing consistently well across both aspects. The direct reconstruction baselines achieve competitive target alignment; however, relying on a single edited guidance image provides limited constraints for non-target regions that are not visible in the selected view, leading to weaker preservation of these regions. In contrast, our TRELLIS-based variant outperforms the compared methods under the TRELLIS setting, while the TRELLIS.2-based implementation achieves the best overall results across all metrics on both benchmarks, demonstrating the compatibility of EditFlow3D with both backbones.

\subsection{Qualitative Evaluation}

\begin{figure}[t]
    \centering
    \includegraphics[width=\columnwidth]{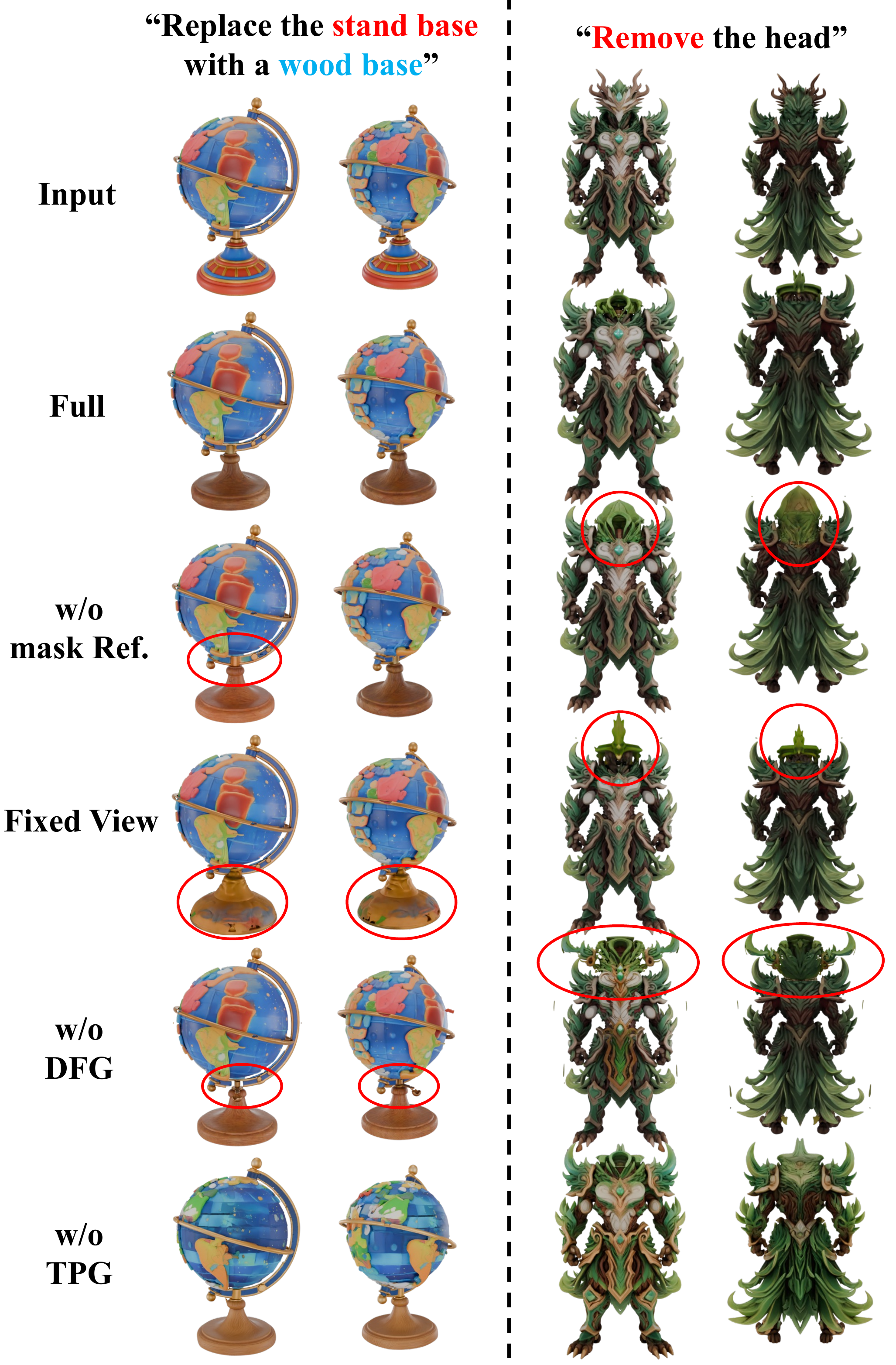}
    \caption{
        Qualitative ablation on EditFlow3D. Each variant changes the control-construction or flow-editing component relative to the full method.
    }
    \label{fig:ablation_component}
\end{figure}

Figure~\ref{fig:show_case} presents qualitative comparisons of local 3D editing. Among the editing baselines, Nano3D introduces noticeable global darkening and appearance shifts, while PartFlow and VoxHammer produce incomplete target modifications, as shown in the second example. Vinedresser3D may also introduce discontinuities near editing boundaries, as illustrated by the first example. The TRELLIS.2 inpainting baseline can generate the requested content, but may alter the geometry and appearance of non-target regions that are insufficiently constrained by the selected view, as highlighted by the red circles. In comparison, EditFlow3D better localizes the requested modification while preserving the structure and appearance of the remaining regions.

\subsection{Ablation Study}
\label{sec:ablation}

\begin{figure}[t]
    \centering
    \includegraphics[width=\columnwidth]{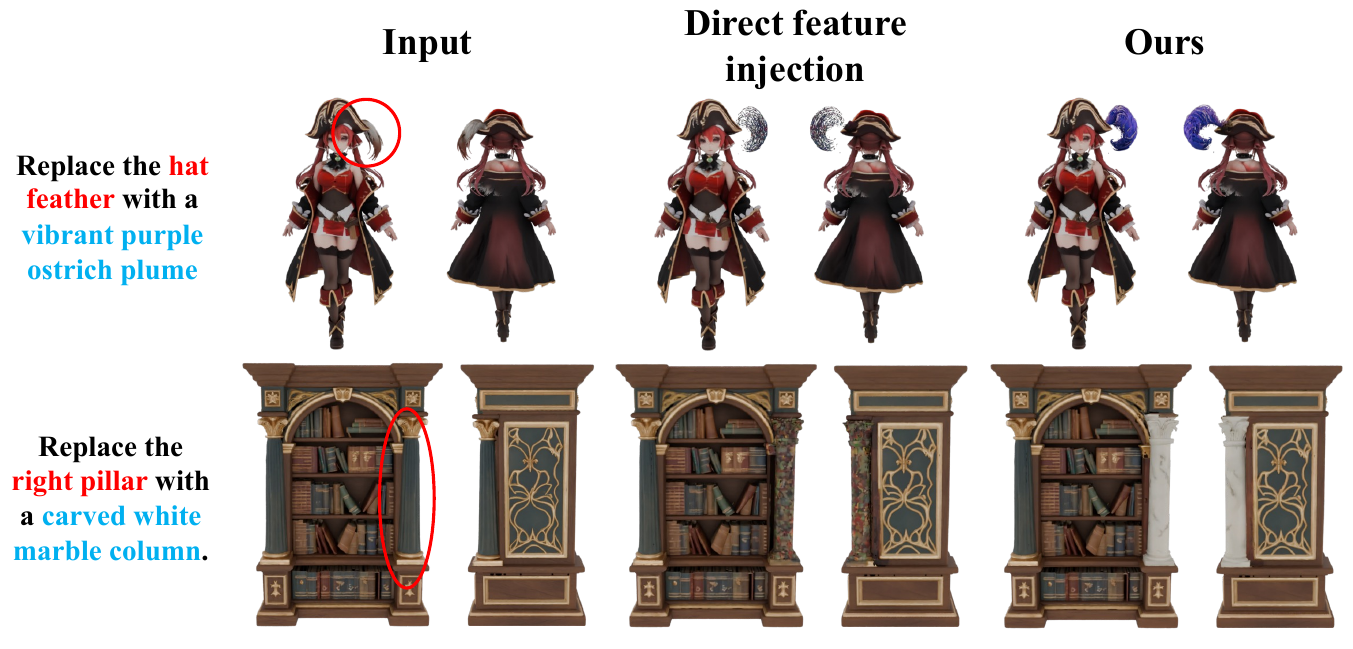}
    \caption{
    Comparison between trajectory preservation guidance and direct feature injection. Direct injection causes geometric fragmentation and texture corruption, whereas TPG better preserves cross-region coherence.
}
    \label{fig:direct_injection}
\end{figure}

Table~\ref{tab:ablation_edit3d} quantitatively evaluates the effects of adaptive view selection, mask refinement, the spatial mask on the differential update, and TPG, while Figure~\ref{fig:ablation_component} provides representative qualitative comparisons of these four components. We further compare TPG with direct feature injection in Figure~\ref{fig:direct_injection}.

\paragraph{Key Component Ablation}
As shown in Table 2 and Figure 4, using a fixed canonical view or only the initial mask degrades both editing accuracy and source preservation, demonstrating the benefits of adaptive view selection and mask refinement. Removing DFG reduces Sim. from 0.318 to 0.287 and substantially worsens all preservation metrics, indicating that DFG is essential for producing the intended target-region modification. In contrast, removing TPG has a smaller effect on Sim., but degrades PSNR, SSIM, LPIPS, and FID to 27.95, 0.846, 0.158, and 67.84, respectively. The qualitative results consistently show pronounced geometry and appearance drift in non-target regions. These results confirm that DFG mainly drives the requested edit, whereas TPG preserves unedited content throughout generation, making the two components complementary.

\paragraph{Direct Feature Injection.}
We compare TPG with direct feature injection, which replaces features in $M_{\mathrm{keep}}$ with source counterparts at each flow step. As shown in Figure~\ref{fig:direct_injection}, hard replacement can make the source features incompatible with the evolving edited region. In the plume-replacement example, it produces fragmented geometry and disconnected structures around the hat. In the pillar-replacement example, it introduces texture corruption and appearance drift in the surrounding cabinet, including non-target regions. With TPG, our model does not overwrite intermediate features. Its soft gradient-based correction preserves the continuity between edited and retained regions while constraining non-target content toward the encoded source state.



\subsection{User Study}
\label{sec:user_study}

In addition to automatic evaluation, we conduct a user study with $40$ participants. We select $10$ cases from each of the five edit types in EditFlow-Bench, for a total of $50$ cases. Each participant evaluates $10$ randomly assigned cases, while a balanced assignment ensures that every case is evaluated by eight distinct participants. For each case, participants are shown the source asset, the editing instruction, and anonymized and randomly ordered results from Nano3D, PartFlow, VoxHammer, Vinedresser3D, and our EditFlow3D (TRELLIS.2). Participants select one preferred result based on each of three criteria: instruction alignment, visual quality, and source preservation. This protocol yields $400$ selections for each criterion. As shown in Table~\ref{tab:user}, participants consistently prefer EditFlow3D for its accurate local edits, high visual quality, and effective preservation of unedited content.

\begin{table}[t]
  \centering
  
  \label{tab:user_study}
  {\small
  \setlength{\tabcolsep}{2pt}
  \begin{tabular}{lccc}
    \toprule
    \textbf{Method}
    & \textbf{\shortstack{Instruction\\Align.}}
    & \textbf{\shortstack{Visual\\Quality}}
    & \textbf{\shortstack{Source\\Preserv.}} \\
    \midrule

    Nano3D~\cite{ye2025nano3d}
    & 15\% & 12\% & 10\% \\

    PartFlow~\cite{weng2026feedforward}
    & 8\% & 9\% & 9\% \\

    VoxHammer~\cite{voxhammer}
    & 15\% & \underline{14\%} & \underline{19\%} \\

    Vinedresser3D~\cite{chi2026vinedresser3d}
    & \underline{19\%} & 8\% & 13\% \\

    \textbf{EditFlow3D (Ours)}
    & \textbf{43\%} & \textbf{57\%} & \textbf{49\%} \\

    \bottomrule
  \end{tabular}
  }
  \caption{
    User preference rates for instruction alignment, visual quality, and source preservation. Bold and underlined values indicate the best and second-best results.
    }
   \label{tab:user}
\end{table}
\section{Conclusion}
\label{sec:conclusion}

We presented \textbf{EditFlow3D}, a training-free framework for controllable local editing of 3D assets. EditFlow3D combines VLM-driven construction of visual guidance and a refined 3D editing mask with localized flow editing in the native 3D representation space of a pretrained generative model. Our results show that spatially masking differential flow updates alone cannot fully preserve non-target regions, as target conditioning may still affect them during generation. To address this issue, step-wise trajectory preservation guidance softly constrains the predicted clean state in non-target regions toward the encoded source state without directly replacing intermediate features. Experiments on our newly constructed EditFlow-Bench and the public Edit3D-Bench demonstrate accurate target-region editing and improved preservation of non-target regions across the evaluated structural and appearance edits, with compatibility across two TRELLIS-family backbones.


\bibliography{aaai2027}

\clearpage
\appendix

\setcounter{secnumdepth}{2}

\numberwithin{figure}{section}
\renewcommand{\thefigure}{\thesection\arabic{figure}}
\numberwithin{table}{section}
\renewcommand{\thetable}{\thesection\arabic{table}}
\numberwithin{equation}{section}
\renewcommand{\theequation}{\thesection\arabic{equation}}
\numberwithin{algorithm}{section}
\renewcommand{\thealgorithm}{\thesection\arabic{algorithm}}

\section*{Supplementary Material}

\noindent\textbf{Organization.}
Appendix~A details the VLM-driven editing-control and 3D-mask construction;
Appendix~B describes the construction and quality control of EditFlow-Bench;
Appendix~C provides additional qualitative results and parameter analysis; and
Appendix~D reports the user-study protocol, limitations, and future work.


\section{VLM-Driven Editing-Control Construction}
\label{app:editing_control}

EditFlow3D converts a source 3D asset and an edit request into a source--target
image pair $(I_{\mathrm{src}}, I_{\mathrm{edit}})$ and a volumetric editing
mask $M_{\mathrm{edit}}$. This appendix provides the implementation details of
this automated control-construction process.

\subsection{VLM Interaction and Control Construction}
\label{app:vlm_interaction}

\subsubsection{Understanding the source asset and instruction.}
For a source asset $\mathcal{A}_{src}$ and an edit request consisting of a text
instruction $q$ and an optional reference image $I_{\mathrm{ref}}$, we first
render $J$ RGB views from a fixed set of cameras
$\mathcal{V}=\{v_j\}_{j=1}^{J}$:
\begin{equation}
    \mathcal{I}_{\mathrm{rgb}}
    = \{I_j^{\mathrm{rgb}}=\mathcal{R}(\mathcal{A}_{src},v_j)\}_{j=1}^{J}.
    \label{eq:supp_rgb_views}
\end{equation}
These cameras cover principal and oblique directions, allowing the VLM to
recognize the complete object and inspect components that may be occluded in a
single view. Camera identifiers and image order are included in the prompt.

Qwen3-VL-8B~\cite{qwen3vl2025} receives $q$,
$\mathcal{I}_{\mathrm{rgb}}$, and, when available, $I_{\mathrm{ref}}$. It
identifies the object, resolves the target component, determines the requested
operation, and ranks the candidate views. Its structured output is
\begin{equation}
    y_v =
    (d_{\mathrm{asset}}, d_{\mathrm{target}}, o,
     \mathcal{V}_{\mathrm{rank}}, c_{\mathrm{view}}),
    \label{eq:supp_vlm_understanding}
\end{equation}
where $d_{\mathrm{asset}}$ is a concise description of the object and its
major components, $d_{\mathrm{target}}$ resolves the noun phrase in $q$ to a
physical component or attachment region, and $o$ denotes one of the five
operations: addition, deletion, replacement, geometry modification, or
appearance modification. $\mathcal{V}_{\mathrm{rank}}$ is an ordered list of
candidate views, while $c_{\mathrm{view}}$ is the view-selection confidence.
This preliminary reasoning is important for relational instructions such as ``the left
pillar,'' ``the front smokestack,'' or ``the upper blue flower,'' whose target
cannot be identified from the noun alone.

\subsubsection{Task-adaptive view selection.}
The first element of $\mathcal{V}_{\mathrm{rank}}$ is selected as $v^*$. In the
text-only setting, the VLM ranks a view highly when (i) the target is visible,
(ii) it is minimally occluded, (iii) its attachment boundary is clear, and
(iv) sufficient context remains to recognize the source asset. Thus,
\begin{equation}
    v^*=\operatorname{First}(\mathcal{V}_{\mathrm{rank}}),
    \qquad I_{\mathrm{src}}=\mathcal{R}(\mathcal{A}_{src},v^*).
    \label{eq:supp_selected_view}
\end{equation}
When a reference image specifies a camera or pose, camera correspondence takes
priority: the selected source view should match its azimuth, elevation, object
orientation, and visible surfaces. The reference image is otherwise treated as
an appearance or structural cue and is not assumed to be pixel-aligned with the
source render.

\subsubsection{Guidance-image generation.}
The selected source view, the original instruction, the resolved target
description, and the optional reference image are passed to FLUX.1 Kontext
[dev]~\cite{labs2025flux1kontextflowmatching}. The image editor is instructed
to modify only the resolved target, preserve the camera and the identity of all
non-target components, and maintain a physically plausible attachment to the
source object. This produces
\begin{equation}
    I_{\mathrm{edit}}=
    \mathcal{E}_{\mathrm{FLUX}}
    (I_{\mathrm{src}},q,d_{\mathrm{target}},I_{\mathrm{ref}}).
    \label{eq:supp_guidance_image}
\end{equation}
The pair $(I_{\mathrm{src}},I_{\mathrm{edit}})$ subsequently defines the source
and target visual conditions $(c_{\mathrm{src}},c_{\mathrm{edit}})$ of the 3D
flow editor.

\begin{figure}[t]
    \centering
    \includegraphics[width=\columnwidth]{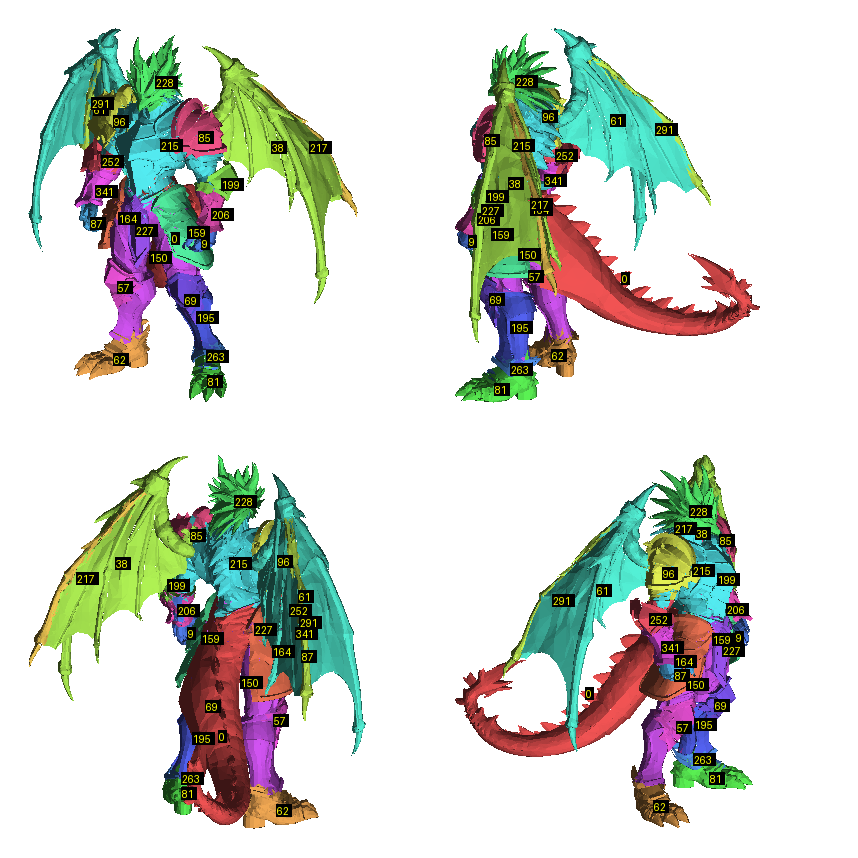}
    \caption{
        Multi-view P3-SAM decomposition with consistent colors and part IDs
        for VLM-based target grounding.
    }
    \label{fig:supp_p3sam_parts}
\end{figure}

\subsubsection{Semantic part grounding.}
In parallel with guidance-image generation, P3-SAM~\cite{ma2025p3} decomposes
$\mathcal{A}_{src}$ into semantic parts. As shown in
Figure~\ref{fig:supp_p3sam_parts}, every part is assigned a persistent integer
ID and a distinct color. We render the labeled mesh from multiple viewpoints to
obtain
\begin{equation}
    \mathcal{I}_{\mathrm{seg}}^{(0)}
    =\{I_{j}^{\mathrm{seg},(0)}\}_{j=1}^{J_s},
    \label{eq:supp_seg_views}
\end{equation}
where the same color and ID denote the same 3D part in every view. Qwen3-VL-8B
is given $q$, $d_{\mathrm{asset}}$, $d_{\mathrm{target}}$, the available part
IDs with their surface-area fractions, and
$\mathcal{I}_{\mathrm{seg}}^{(0)}$. It returns
\begin{equation}
    y_p=
    (o,\mathcal{K}_{\mathrm{VLM}},
     \{c_k\}_{k\in\mathcal{K}_{\mathrm{VLM}}},c_{\mathrm{mask}},r),
    \label{eq:supp_grounding_output}
\end{equation}
where $\mathcal{K}_{\mathrm{VLM}}$ contains every P3-SAM ID belonging to the
target, $c_k$ and $c_{\mathrm{mask}}$ are diagnostic confidences, and $r$ is a
short rationale. Multi-part unions are allowed when a semantic target is split
into several segments or contains repeated instances. The IDs are validated
against the available P3-SAM labels before mask construction. The confidence
values are recorded for auditing but do not change the equal-weight geometric
fusion described below. The complete construction of the visual guidance and
initial target IDs is summarized in Algorithm~\ref{alg:supp_vlm_control}.

\begin{algorithm}[t]
\caption{VLM interaction for editing-control acquisition}
\label{alg:supp_vlm_control}
\begin{algorithmic}[1]
\REQUIRE Source asset $\mathcal{A}_{src}$, instruction $q$, optional reference image
         $I_{\mathrm{ref}}$, source cameras $\mathcal{V}$, segmentation cameras
         $\mathcal{V}_s$
\ENSURE $I_{\mathrm{src}}$, $I_{\mathrm{edit}}$, operation $o$, target IDs
        $\mathcal{K}_{\mathrm{VLM}}$
\STATE Render RGB views $\mathcal{I}_{\mathrm{rgb}}$ of $\mathcal{A}_{src}$ from $\mathcal{V}$
\STATE Query Qwen3-VL-8B with $(q,\mathcal{I}_{\mathrm{rgb}},I_{\mathrm{ref}})$
\STATE Parse asset description $d_{\mathrm{asset}}$, target description
       $d_{\mathrm{target}}$, operation $o$, and ranked views
       $\mathcal{V}_{\mathrm{rank}}$
\STATE Select $v^*\leftarrow\mathcal{V}_{\mathrm{rank}}[1]$ and set
       $I_{\mathrm{src}}\leftarrow\mathcal{R}(\mathcal{A}_{src},v^*)$
\STATE Generate $I_{\mathrm{edit}}\leftarrow
       \mathcal{E}_{\mathrm{FLUX}}(I_{\mathrm{src}},q,
       d_{\mathrm{target}},I_{\mathrm{ref}})$
\STATE Run P3-SAM on $\mathcal{A}_{src}$ and assign persistent colors and IDs
\STATE Render the P3-SAM ID maps $\mathcal{I}_{\mathrm{seg}}^{(0)}$ from
       $\mathcal{V}_s$
\STATE Query Qwen3-VL-8B with $(q,d_{\mathrm{asset}},d_{\mathrm{target}},
       \mathcal{I}_{\mathrm{seg}}^{(0)})$
\STATE Parse and validate all returned target IDs
       $\mathcal{K}_{\mathrm{VLM}}$
\RETURN $(I_{\mathrm{src}},I_{\mathrm{edit}},o,
         \mathcal{K}_{\mathrm{VLM}})$
\end{algorithmic}
\end{algorithm}

\subsection{Multi-Model 3D Mask Refinement}
\label{app:mask_refinement}

\subsubsection{Common surface points and initial mask.}
Different segmentation models may internally remesh or resample the asset.
Consequently, face indices are not assumed to correspond across models. We
normalize $\mathcal{A}_{src}$ into a shared coordinate frame and uniformly sample
$N=100{,}000$ surface points,
\begin{equation}
    \mathcal{P}=\{(p_i,n_i,w_i)\}_{i=1}^{N},
    \label{eq:supp_common_domain}
\end{equation}
where $n_i$ is the surface normal and $w_i$ is the area weight. Every
segmentation is transferred to $\mathcal{P}$ by nearest-neighbor matching in
normalized 3D space. For any $S\subseteq\mathcal{P}$, its area is
$A(S)=\sum_{p_i\in S}w_i$.

Let $S_k^{(0)}$ be the P3-SAM part with ID $k$. The VLM-selected union is
\begin{equation}
    M^{(0)}=\bigcup_{k\in\mathcal{K}_{\mathrm{VLM}}}S_k^{(0)}.
    \label{eq:supp_initial_mask}
\end{equation}
P3-SAM supplies the semantic identity of the target, whereas the auxiliary
models are used to correct geometric leakage and missing regions near part
junctions.

\subsubsection{Multi-granularity auxiliary matching.}
Let $m\in\{1,2\}$ denote PartField~\cite{liu2025partfield} and
S2AM3D~\cite{su2026s2am3d}, respectively. Both models are
processed identically. Model $m$ produces candidate parts
$S_k^{(m)}\subseteq\mathcal{P}$ at multiple granularities, and
$g_k^{(m)}$ records the granularity of candidate $k$. We compute
\begin{align}
    P_k^{(m)}
    &=\frac{A(S_k^{(m)}\cap M^{(0)})}{A(S_k^{(m)})},
    \label{eq:supp_purity}
\end{align}
Purity rejects candidates that contain substantial unrelated geometry. For each granularity $g$, the candidate pool is
\begin{equation}
    \mathcal{Q}_{m,g}=\{k\mid g_k^{(m)}=g,
    \;P_k^{(m)}\geq\tau_P\}.
    \label{eq:supp_candidate_pool}
\end{equation}

Starting from $U_{m,g}=\varnothing$, we greedily add the candidate producing
the largest positive increase in area-weighted $F_1$ with respect to
$M^{(0)}$. For a union $U$, define
\begin{align}
    \operatorname{Prec}(U)&=\frac{A(U\cap M^{(0)})}{A(U)},\\
    \operatorname{Rec}(U)&=\frac{A(U\cap M^{(0)})}{A(M^{(0)})},\\
    F_1(U,M^{(0)})&=
    \frac{2\operatorname{Prec}(U)\operatorname{Rec}(U)}
    {\operatorname{Prec}(U)+\operatorname{Rec}(U)}.
    \label{eq:supp_area_f1}
\end{align}
We define $F_1(\varnothing,M^{(0)})=0$.
The search terminates when no candidate increases $F_1$ or when at most
$K_{\max}$ candidates have been selected. The best granularity is then
selected independently for each auxiliary model:
\begin{equation}
    g_m^*=\arg\max_g F_1(U_{m,g},M^{(0)}),
    \qquad M^{(m)}=U_{m,g_m^*}.
    \label{eq:supp_best_granularity}
\end{equation}
Each accepted candidate is retained in its entirety, rather than clipped to
$M^{(0)}$, allowing PartField and S2AM3D to recover geometry omitted by
P3-SAM.

\subsubsection{Equal-weight consensus.}
The three masks are fused with equal weight. For each canonical point,
\begin{equation}
    s_i=\frac{1}{3}\sum_{m=0}^{2}
    \mathbb{I}[p_i\in M^{(m)}].
    \label{eq:supp_equal_vote}
\end{equation}
Points supported by at least two models form reliable foreground
$\mathcal{P}^{+}=\{p_i\mid s_i\geq2/3\}$. Points supported by none form
reliable background $\mathcal{P}^{-}=\{p_i\mid s_i=0\}$. The remaining
one-model points form the uncertain set. This construction
prevents a single auxiliary partition from overriding the semantic target,
while exposing model disagreement explicitly as a narrow refinement region.

\subsubsection{Normal-aware mesh-graph refinement.}
We transfer the point-wise votes to the source mesh faces. For face $f$, let
$\bar{s}_f$ be the area-weighted mean vote of its sampled points. A face is a
reliable foreground or background anchor when at least half of its sampled
area belongs to $\mathcal{P}^{+}$ or $\mathcal{P}^{-}$, respectively. Faces
without sufficient evidence remain uncertain.

We construct the face graph $G=(\mathcal{F},\mathcal{E})$, where two faces are
adjacent when they share a mesh edge. Reliable labels are fixed, and uncertain
face labels $\ell_f\in\{0,1\}$ minimize
\begin{equation}
\begin{split}
    E(\boldsymbol{\ell})=
    &\sum_{f\in\mathcal{F}^{?}}
    \left[-\ell_f\log\bar{s}_f-(1-\ell_f)\log(1-\bar{s}_f)\right]\\
    &+\lambda_{\mathrm{G}}\sum_{(f,h)\in\mathcal{E}}
    \left[\max(\langle n_f,n_h\rangle,0)\right]^{\eta}
    \mathbb{I}[\ell_f\neq \ell_h],
    \label{eq:supp_graph_energy}
\end{split}
\end{equation}
where $n_f$ is the face normal. The pairwise term strongly encourages
consistent labels across smooth surfaces, but becomes weak across sharp normal
discontinuities, allowing the boundary to stop at geometric part junctions.
We optimize Eq.~\eqref{eq:supp_graph_energy} by iterated conditional modes
(ICM) while keeping the reliable anchors fixed.

After graph optimization, the mesh faces corresponding to $M^{(0)}$ define the
target region. Foreground components disconnected from this region
are rejected, and remaining foreground components whose area is less than
$\alpha_{\mathrm{comp}}$ of the selected area are removed. A background
component is filled only when it is enclosed by foreground through shared-edge
adjacency and its area is below $\alpha_{\mathrm{hole}}$ of the complete asset
area. Thus, connectivity to the target suppresses unrelated candidate regions,
while the normal-aware pairwise term places the refined boundary at geometric
part junctions. Requiring an actual enclosing boundary also prevents
disconnected mesh components from being mistaken for holes. The resulting
surface mask is $M_{\mathrm{ref}}$.

\subsubsection{Operation-aware volumetric mask.}
We map $M_{\mathrm{ref}}$ to the native sparse coordinates of the 3D
generative model. For deletion, we use a small face-ring margin to cover the
complete removable component. For replacement and geometric modification, a
larger bounded neighborhood gives the generator room to change the target
silhouette and attachment. Appearance modification uses the selected surface
region with a narrow boundary margin because it does not create new geometry
outside the current surface. For addition, the VLM-selected component acts as
an attachment anchor; nearby unoccupied sparse coordinates outside the source
surface are marked editable together with a narrow interface margin. We write
this operation-specific mask construction as
\begin{equation}
    M_{\mathrm{edit}}=\mathcal{D}_{o}(M_{\mathrm{ref}}),
    \qquad M_{\mathrm{keep}}=1-M_{\mathrm{edit}}.
    \label{eq:supp_edit_keep_masks}
\end{equation}
$M_{\mathrm{ref}}$ is first voxelized at the native asset resolution and then
mapped to the sparse-structure resolution. Occupied non-target voxels remain in
$M_{\mathrm{keep}}$, whereas editable empty coordinates permit the creation of
new geometry. Algorithm~\ref{alg:supp_mask_refinement} summarizes the complete
process from candidate matching to the construction of $M_{\mathrm{edit}}$ and
$M_{\mathrm{keep}}$.

\begin{algorithm}[t]
\caption{Multi-model 3D mask matching, voting, and refinement}
\label{alg:supp_mask_refinement}
\scriptsize
\begin{algorithmic}[1]
\REQUIRE Source asset $\mathcal{A}_{src}$, VLM-selected P3-SAM IDs
         $\mathcal{K}_{\mathrm{VLM}}$, PartField partitions,
         S2AM3D partitions, operation $o$
\ENSURE Refined surface mask $M_{\mathrm{ref}}$, volumetric masks
        $(M_{\mathrm{edit}},M_{\mathrm{keep}})$
\STATE Normalize $\mathcal{A}_{src}$ and sample a common set of surface points
       $\mathcal{P}$ with area weights
\STATE Transfer P3-SAM, PartField, and S2AM3D labels to $\mathcal{P}$
\STATE $M^{(0)}\leftarrow\bigcup_{k\in\mathcal{K}_{\mathrm{VLM}}}S_k^{(0)}$
\FOR{$m\in\{\mathrm{PartField},\mathrm{S2AM3D}\}$}
    \FOR{each granularity $g$ of model $m$}
        \STATE $\mathcal{Q}_{m,g}\leftarrow
        \{k\mid g_k^{(m)}=g,\;P_k^{(m)}\geq\tau_P\}$
        \STATE $U\leftarrow\varnothing$, $b\leftarrow0$, $r\leftarrow0$
        \WHILE{$\mathcal{Q}_{m,g}\neq\varnothing$ and $r<K_{\max}$}
            \STATE $k^*\leftarrow\arg\max_{k\in\mathcal{Q}_{m,g}}
            F_1(U\cup S_k^{(m)},M^{(0)})$
            \STATE $b^*\leftarrow F_1(U\cup S_{k^*}^{(m)},M^{(0)})$
            \IF{$b^*\leq b+\epsilon_{\mathrm{gain}}$}
                \STATE \textbf{break}
            \ENDIF
            \STATE $U\leftarrow U\cup S_{k^*}^{(m)}$,
            $b\leftarrow b^*$, $r\leftarrow r+1$
            \STATE Remove $k^*$ from $\mathcal{Q}_{m,g}$
        \ENDWHILE
        \STATE Store $U_{m,g}\leftarrow U$
    \ENDFOR
    \STATE $g_m^*\leftarrow\arg\max_g F_1(U_{m,g},M^{(0)})$
    \STATE $M^{(m)}\leftarrow U_{m,g_m^*}$
\ENDFOR
\FOR{each canonical point $p_i$}
    \STATE $s_i\leftarrow\frac{1}{3}\sum_{m=0}^{2}
    \mathbb{I}[p_i\in M^{(m)}]$
    \STATE Assign $p_i$ to reliable foreground if $s_i\geq2/3$,
    reliable background if $s_i=0$, and uncertain otherwise
\ENDFOR
\STATE Transfer point votes and reliability states to mesh faces
\STATE Construct the shared-edge face graph and identify the faces corresponding
       to $M^{(0)}$
\STATE Initialize uncertain faces with $\ell_f\leftarrow\mathbb{I}[\bar{s}_f\geq1/2]$
\STATE Optimize uncertain labels by ICM using
       Eq.~\eqref{eq:supp_graph_energy}
\STATE Reject foreground components disconnected from the target region
\STATE Remove small foreground components and fill only small enclosed holes
\STATE $M_{\mathrm{ref}}\leftarrow\{f\mid \ell_f=1\}$
\STATE Apply the operation-specific surface margin or empty-space expansion
       $M_{\mathrm{edit}}\leftarrow\mathcal{D}_{o}(M_{\mathrm{ref}})$
\STATE Voxelize $M_{\mathrm{edit}}$ and set
       $M_{\mathrm{keep}}\leftarrow1-M_{\mathrm{edit}}$
\RETURN $(M_{\mathrm{ref}},M_{\mathrm{edit}},M_{\mathrm{keep}})$
\end{algorithmic}
\end{algorithm}

\subsubsection{Implementation parameters.}
Unless otherwise stated, we use $N=100{,}000$ canonical surface samples,
$\tau_P=0.6$, $K_{\max}=8$, and
$\epsilon_{\mathrm{gain}}=10^{-6}$. The mesh-graph optimizer uses
$\lambda_{\mathrm{G}}=1$, $\eta=4$, and at most $T_{\max}=12$ ICM iterations. We set
$\alpha_{\mathrm{comp}}=0.005$ and
$\alpha_{\mathrm{hole}}=0.02$. The deletion mask is expanded by one
shared-edge face ring, while replacement and geometric-modification masks are
expanded by four rings. Appearance modification uses one boundary ring.
PartField and S2AM3D contribute one matched mask each, and their votes have
exactly the same weight as the P3-SAM mask.


\section{EditFlow-Bench Construction}
\label{app:benchmark_construction}

\subsection{Source Asset Collection}

EditFlow-Bench contains 100 source assets: 30 real-scanned objects from Google
Scanned Objects (GSO)~\cite{downs2022scanned}, 30 artist-created assets
corresponding to real-world products from Amazon Berkeley Objects
(ABO)~\cite{Collins_2022_ABO}, 20 assets generated by
TRELLIS.2~\cite{xiang2025native}, and 20 generated by Hunyuan3D
2.1~\cite{hunyuan3d-2.1}. This composition covers scanned, manually modeled,
and generatively synthesized content, reducing dependence on any single asset
source.

We retain object-centric assets with complete geometry and appearance, a
recognizable identity, and locally editable components or attachment regions.
Assets with severe reconstruction artifacts, incomplete surfaces or materials,
degenerate geometry that prevents stable rendering or segmentation, or no
spatially identifiable target are excluded, as are duplicate and
near-duplicate instances. Each accepted asset is centered and isotropically
scaled to a shared canonical bounding volume while preserving its relative
geometry, topology, and materials. The normalized asset is used consistently
for rendering, segmentation, mask construction, editing, and evaluation.

\subsection{Editing Case Construction}

Each asset is associated with two distinct editing cases, resulting in 200
cases evenly divided among addition, deletion, replacement, geometry
modification, and appearance modification, with 40 cases per operation.
Addition introduces a component at a defined attachment region; deletion
removes an existing component; replacement substitutes a selected component;
geometry modification changes its shape, scale, or structure; and appearance
modification changes local color, texture, or material attributes without
requiring a structural change.

Candidate instructions are generated using Gemini 3
Pro~\cite{gemini3pro2026}. We retain only instructions that identify a single
local target or an unambiguous union of related parts, specify a visually
verifiable change, and leave the intended state of non-target content
unchanged. Relational descriptions are allowed only when the target can be
resolved consistently from the source views, and the two cases for an asset
must request distinct outcomes.

For each instruction, our automated workflow selects a source view and FLUX.1
Kontext [dev]~\cite{labs2025flux1kontextflowmatching} generates the guidance
image. A case is accepted only when the image depicts the requested local
change, preserves the camera and non-target identity, and contains no severe
artifact that makes the desired 3D result ambiguous. Otherwise, the complete
candidate is discarded and regenerated. This process continues until every
asset has two valid cases and each operation contains exactly 40 cases.

\subsection{Automatic Mask Construction and Quality Control}

Every EditFlow-Bench mask is generated by the workflow in
Appendix~\ref{app:mask_refinement}. It constructs an initial semantic mask from
VLM-selected P3-SAM part IDs, matches and fuses the PartField and S2AM3D
partitions, applies target-connected normal-aware boundary refinement, and
converts the result into an operation-aware volumetric mask. No manual point,
box, part ID, surface region, or refinement parameter is provided.

Human review is used only as a binary quality check after automatic mask
construction and before any editing result is generated. Seven reviewers
conducted this audit, requiring approximately 30 person-hours in total.
Reviewers inspect the mask from multiple views and verify that it covers the
intended target and necessary attachment region without substantial unrelated
geometry. For addition and structural replacement, they also check the
editable volumetric support around the attachment. Reviewers cannot paint,
erase, resize, relabel, or otherwise modify a mask or its automatically
selected part IDs. A failed case is discarded and regenerated from the
beginning; it is never manually repaired. Thus, EditFlow-Bench masks are
automatically generated and only human-verified.

This differs from Edit3D-Bench, where mask-based methods use the provided
human-annotated oracle masks. On EditFlow-Bench, all mask-based methods receive
the same accepted automatic masks. Because the pass--fail audit occurs before
method execution and never examines editing results, it does not favor a
particular editing method.


\section{Additional Results and Analysis}
\label{app:additional_results}

\subsection{Qualitative Comparison Using TRELLIS}

\begin{figure}[t]
    \centering
    \includegraphics[width=\columnwidth]{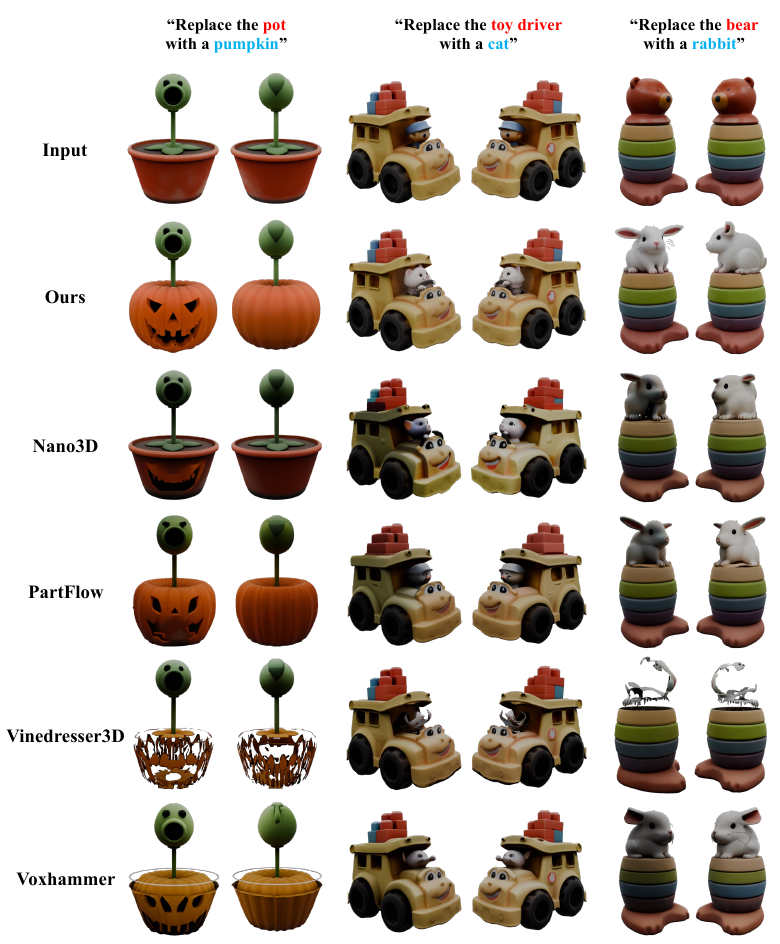}
    \caption{
        Qualitative comparison using TRELLIS. All methods use the same source asset and
        edited guidance image for each case.
    }
    \label{fig:supp_trellis_comparison}
\end{figure}

Figure~\ref{fig:supp_trellis_comparison} presents a qualitative comparison
using the TRELLIS backbone~\cite{trellis}. We compare EditFlow3D with
Nano3D~\cite{ye2025nano3d}, PartFlow~\cite{weng2026feedforward},
Vinedresser3D~\cite{chi2026vinedresser3d}, and
VoxHammer~\cite{voxhammer}. All methods receive the same source asset and
edited guidance image for each case. Mask-based methods additionally follow
the same benchmark-specific mask protocol described in the main paper and
Appendix~\ref{app:benchmark_construction}. The comparison shows that EditFlow3D
retains its advantage in localized editing and source preservation when using
TRELLIS.


\subsection{Additional Editing Results}

As shown in Figure~\ref{fig:supp_trellis_comparison}, the compared methods may
produce incomplete target modifications, introduce global appearance changes,
or create discontinuities near the editing boundary. In contrast, EditFlow3D
more consistently confines the requested change to the intended component and
retains the structure and appearance of surrounding regions. Since all results
in this comparison use the same TRELLIS backbone, the observed
improvement cannot be attributed solely to the stronger generative capacity of
TRELLIS.2. This comparison is also consistent with the TRELLIS-based
quantitative results reported in the main paper.

\begin{figure}[t]
    \centering
    \includegraphics[width=\columnwidth]{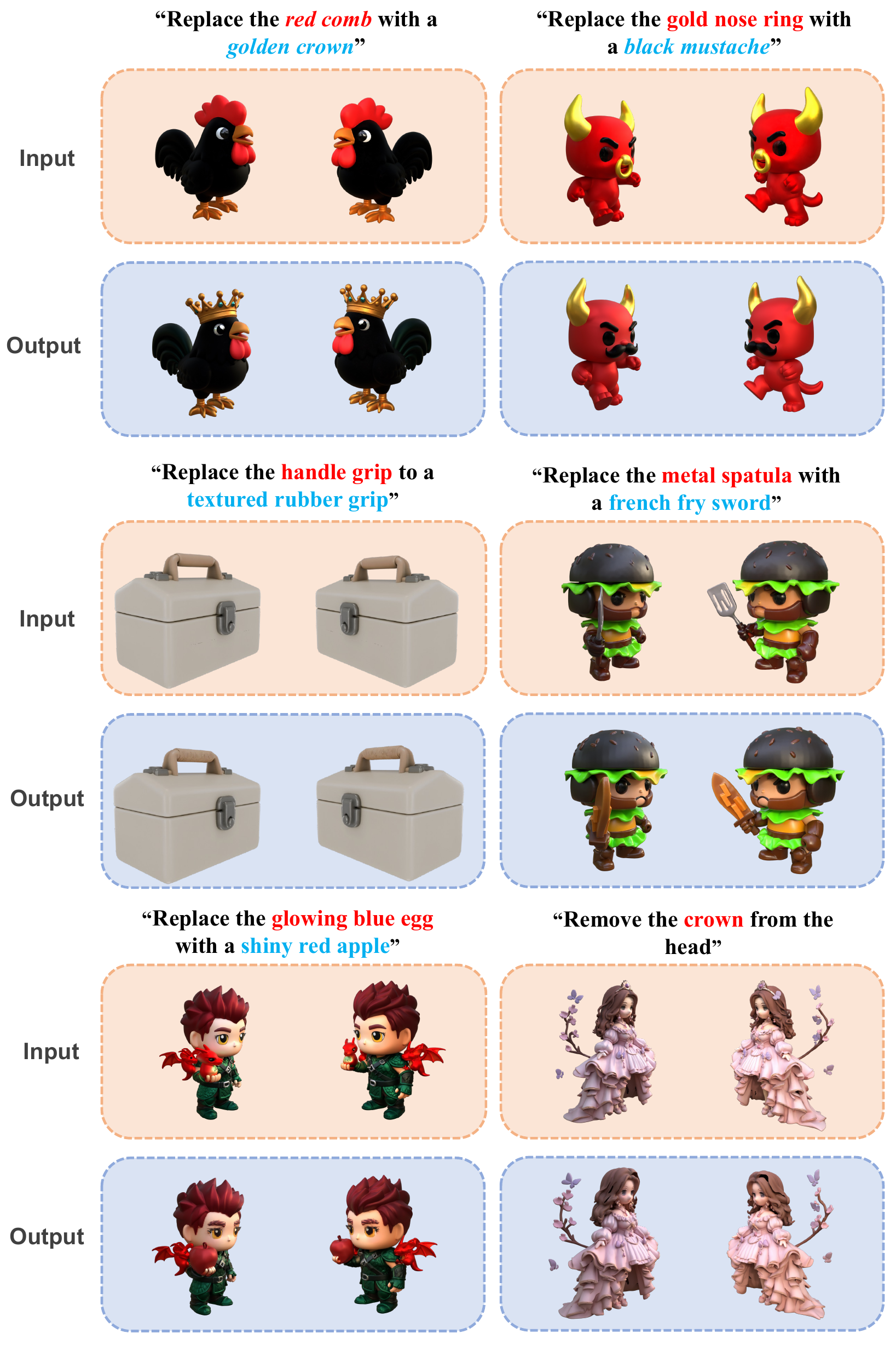}
    \caption{
        Additional local editing results of EditFlow3D. Each case shows two
        views of the source asset and the edited output.
    }
    \label{fig:supp_more_cases}
\end{figure}

Figure~\ref{fig:supp_more_cases} presents additional local editing results on
assets with different geometric complexity, target sizes, and attachment
relationships. The examples include replacing compact components, replacing
larger structures whose silhouettes differ substantially from the source
parts, and deleting an existing component. For each case, two rendered views
are shown to reveal both the requested local change and its consistency across
viewpoints.

These results further demonstrate that EditFlow3D is not limited to a
particular object category or target scale. The edited components remain
structurally attached to the source objects, while non-target shapes,
materials, and fine appearance details remain consistent across the shown
views. The examples also illustrate that the same editing formulation can
handle both removal and substantial part replacement without requiring
case-specific optimization or training.

\subsection{Effect of the Trajectory Preservation Strength}

\begin{figure}[t]
    \centering
    \includegraphics[width=\columnwidth]{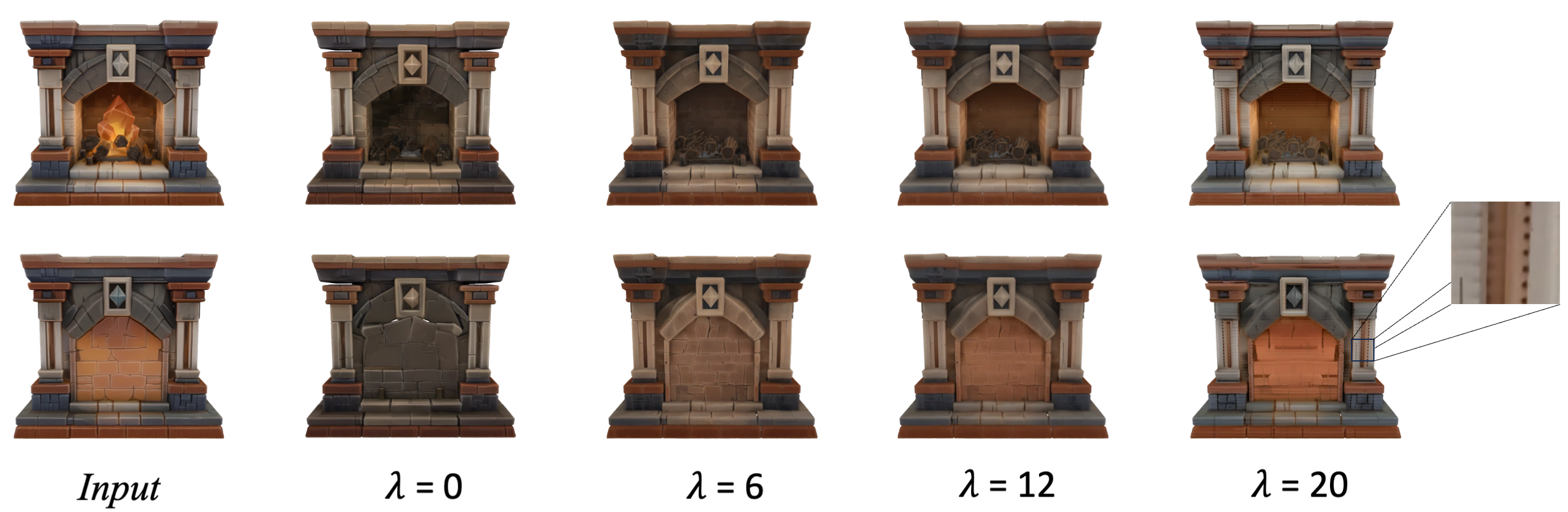}
    \caption{
        Effect of the TPG preservation strength $\lambda$ for the instruction
        ``Delete the fire in the stove.'' A moderate value of $\lambda=12$ provides the best
        balance between the requested edit and preservation of the source
        asset.
    }
    \label{fig:supp_lambda_ablation}
\end{figure}

The preservation strength $\lambda$ controls the gradient correction introduced
by TPG in the non-target region. We analyze
$\lambda\in\{0,6,12,20\}$ using the instruction ``Delete the fire in the
stove.'' As shown in Figure~\ref{fig:supp_lambda_ablation}, setting
$\lambda=0$ removes the trajectory-preservation constraint. Although the target
edit can still be performed, the geometry and appearance of the surrounding
fireplace drift noticeably from the source asset. A moderate constraint with
$\lambda=6$ reduces this drift but does not fully recover the source structure
and material consistency.

With $\lambda=12$, non-target geometry and appearance remain close to the
source asset while the fire is successfully removed, yielding a coherent
transition around the editing boundary. Increasing the strength to
$\lambda=20$ over-constrains the feature trajectory. The stronger correction
begins to interfere with the evolution of features near the target boundary,
leading to residual appearance changes and visible boundary artifacts. These
results illustrate the expected trade-off: a weak constraint provides
insufficient source preservation, whereas an overly strong constraint can
suppress or disturb the requested local edit. We therefore use $\lambda=12$ in
all other experiments.


\section{User Study and Discussion}
\label{app:discussion}

\subsection{User Study Details}

We recruit 40 participants and sample 50 cases from EditFlow-Bench, with 10
cases from each of the five edit types. Each participant evaluates 10 randomly
assigned cases, and a balanced assignment ensures that every case is evaluated
by eight distinct participants. For each case, participants are shown the
source asset, the editing instruction, and anonymized, randomly ordered results
from Nano3D, PartFlow, VoxHammer, Vinedresser3D, and EditFlow3D using
TRELLIS.2. All results are rendered from the same viewpoints.

Participants select one preferred result separately for instruction alignment,
visual quality, and source preservation. These criteria assess whether the
target edit follows the instruction, whether the result is visually coherent
and artifact-free, and whether non-target geometry and appearance remain
consistent with the source asset, respectively. The balanced protocol yields
400 selections for each criterion. EditFlow3D receives preference rates of
43\%, 57\%, and 49\% for the three criteria, respectively, ranking first in
all cases.

\subsection{Limitations and Future Work}

EditFlow3D depends on the quality of its automatically constructed visual
guidance and 3D mask; incorrect view selection, ambiguous part grounding, or
segmentation errors may therefore propagate to the final edit, particularly
for thin or heavily occluded components. Future work could jointly assess
guidance and mask confidence and automatically retry failed control
construction. In addition, TPG currently uses a fixed preservation strength
$\lambda$. A spatially and temporally adaptive preservation constraint, together with evaluation on broader 3D representations, may provide a better balance between editability
and source preservation.

\end{document}